\documentclass[runningheads]{llncs}
\usepackage[T1]{fontenc}
\usepackage{graphicx}
\usepackage{tabularx}
\usepackage{cite}
\usepackage{amsmath,amssymb,amsfonts}
\usepackage{algorithmic}
\usepackage{graphicx}
\usepackage{textcomp}
\usepackage{xcolor}
\usepackage{academicons}
\usepackage[misc, geometry]{ifsym}
\usepackage{siunitx}
\usepackage{booktabs}
\usepackage{multirow}
\usepackage{svg}
\usepackage{amsmath}
\usepackage{wrapfig}
\usepackage{sidecap}
\usepackage{multirow}
\usepackage{wrapfig}

\usepackage{listings}

\usepackage{float} 

\newbox{\myorcidaffilbox}
\sbox{\myorcidaffilbox}{\large\includegraphics[height=1.25ex]{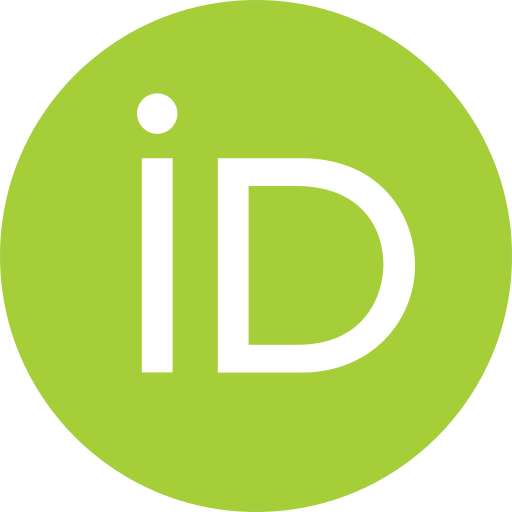}}
\newcommand{\orcidaffil}[1]{%
  \href{https://orcid.org/#1}{\usebox{\myorcidaffilbox}}}

\newcommand{\comment}[1]{\ignorespaces}

\usepackage{hyperref}
\usepackage{color}

\def\BibTeX{{\rm B\kern-.05em{\sc i\kern-.025em b}\kern-.08em
    T\kern-.1667em\lower.7ex\hbox{E}\kern-.125emX}}

\begin{document}
\title{PB\&J: Peanut Butter and Joints for Damped Articulation}
%
%
\author{Avery S. Williamson\inst{1}$^{+}$\orcidaffil{0009-0002-3945-2816} \and
Michael J. Bennington\inst{1}$^{+}$\orcidaffil{0000-0001-7940-6297}\and
Ravesh Sukhnandan\inst{1}$^{+}$\orcidaffil{0000-0001-5858-9610} 
\and Mrinali Nakhre\inst{1}
\and Yuemin Mao\inst{5}
\and Victoria A. Webster-Wood\inst{1,2,3,4,5}\orcidaffil{0000-0001-6638-2687}}
%
%
\institute{Depts. of $^1$Mechanical Engineering, $^2$Biomedical Engineering, and $^3$Material Science, $^4$McGowan Institute for Regenerative Medicine, $^5$Robotics Institute, Carnegie Mellon University, Pittsburgh, PA, USA \\
$^{+}$These authors contributed equally to the work.\\
\Letter \email{vwebster@andrew.cmu.edu}}
\maketitle              
\begin{abstract}
Many bioinspired robots mimic the rigid articulated joint structure of the human hand for grasping tasks, but experience high-frequency mechanical perturbations that can destabilize the system and negatively affect precision without a high-frequency controller. Despite having bandwidth-limited controllers that experience time delays between sensing and actuation, biological systems can respond successfully to and mitigate these high-frequency perturbations. Human joints include damping and stiffness that many rigid articulated bioinspired hand robots lack. To enable researchers to explore the effects of joint viscoelasticity in joint control, we developed a human-hand-inspired grasping robot with viscoelastic structures that utilizes accessible and bioderived materials to reduce the economic and environmental impact of prototyping novel robotic systems. We demonstrate that an elastic element at the finger joints is necessary to achieve concurrent flexion, which enables secure grasping of spherical objects. To significantly damp the manufactured finger joints, we modeled, manufactured, and characterized rotary dampers using peanut butter as an organic analog joint working fluid. Finally, we demonstrated that a real-time position-based controller could be used to successfully catch a lightweight falling ball. 
We developed this open-source, low-cost grasping platform that abstracts the morphological and mechanical properties of the human hand to enable researchers to explore questions about biomechanics \textit{in roboto} that would otherwise be difficult to test in simulation or modeling. 

\end{abstract}

\section{Introduction}

Adaptability in the face of dynamic and unknown environments is vital to any system we design to interact with our human world. To date, this remains a significant hurdle in the development of robotic systems, with many robotic systems failing when brought out of the lab into a new environment \cite{nishikawa_neuromechanics_2007,guizzo_darpa_nodate}. On the other hand, animals routinely perform these tasks, locomoting in unknown and dynamic environments and manipulating objects for the sake of hunting or survival, from a very young age \cite{nishikawa_neuromechanics_2007,holme_nielsen_rapid_2018}. 

Part of the difficulty in these tasks arises from the mismatch between the range of frequencies used for movement and the range of frequencies experienced during perturbations. The majority of human and large animal actions range from 0 Hz (standing, hanging, etc.) to mid-10 Hz \cite{Lakie2012, Vernooij2015, Sutton, Ingram2008}, and for robots with which we hope to cohabitate, this same range of motions would be required. However, these robots also need to behaviorally respond to a larger range of mechanical perturbations \cite{Dong2021, Liu2007}, including impulsive collisions (falling, hitting objects, stopping objects, etc.) with forcing frequencies up to 1000s of Hz. 

Current robots overcome this mismatch by utilizing high-bandwidth sensory systems/controllers and increasingly complex control paradigms. To ensure a wide range of disturbances are sensed and compensated, robots have sensory systems running in the kilohertz range \cite{Namiki2003}, with processing centers running ever faster to process that level of information. Controllers responding to these sensory inputs increasingly include computationally expensive real-time physics models\cite{Guizzo}, multilayer/hybrid controllers \cite{Burridge}, and black-box machine learning algorithms \cite{Tuong} that can be economically costly and energy intensive. In contrast, animals that can achieve the same adaptive behaviors are controlled by neural circuits, the base units of which (the neurons) exhibit a biophysically limited computation rate \cite{Fundamentals}. The minimum time between sequential neuron firings (refractory period) can range from 1-10 milliseconds \cite{Fundamentals}, limiting the spike rates to 100-1000 Hz. Furthermore, the decreased conductivity of nerves compared to metal wires limits signal transmission rates to 0.1-100 m / s \cite{Fundamentals}, which can cause up to 100 ms control loop delays. Despite these transmission limitations, animals can recover from and adapt to extreme mechanical perturbations to perform precise locomotive and manipulation tasks.

What these animal systems have that is often neglected (or intentionally removed) in robot systems is viscoelasticity. In animals, this occurs passively from the fluid-solid composite structure of joints \cite{Lakie2012, Vernooij2015, Sutton, Ingram2008, Murray1994}, and both actively and passively from velocity-dependent force generation of skeletal muscle \cite{Murray1994}. In the case of the human hand, passive damping occurs in the joint capsules between different segments of the fingers, and active damping occurs in both the intrinsic and extrinsic muscles that articulate the fingers\cite{Kamper2002}. Although damping is typically seen as a source of energy loss in engineered systems, it could provide key morphological computational resources for biological systems, serving as low-pass force filters. This low-pass filtering of mechanical perturbations could increase stability in the face of disturbances, and the phase shifts associated with filters could compensate for the control phase shifts introduced by limited conduction rates. 

Replicating this filtering in robotic systems using active and passive damping could provide an alternative approach to traditional computationally and financially expensive control methods. Toward investigating the role of viscoelastic joints in precision and adaptive behaviors in the face of high-frequency perturbations, we present the design, modeling, and characterization of an open-source biomimetic grasper based on the human hand. We utilize low-cost rapid prototyping tools and readily available materials and electronic components that prioritize accessibility, modularity, and reliance, when possible, on biologically derived materials that aim to minimize economic and environmental impact. This grasper incorporates both joint elasticity and damping mechanisms that could be tuned to achieve different degrees of joint viscoelasticity. With this \textit{in roboto} platform, roboticists and biologists could investigate the role of viscoelasticity in human-like grasping. 



\section{Design and Modeling}
\subsection{Design and CAD Modeling of Hand Prototype}

\begin{figure}[!t]
    \centering
    \includegraphics[width=\linewidth]{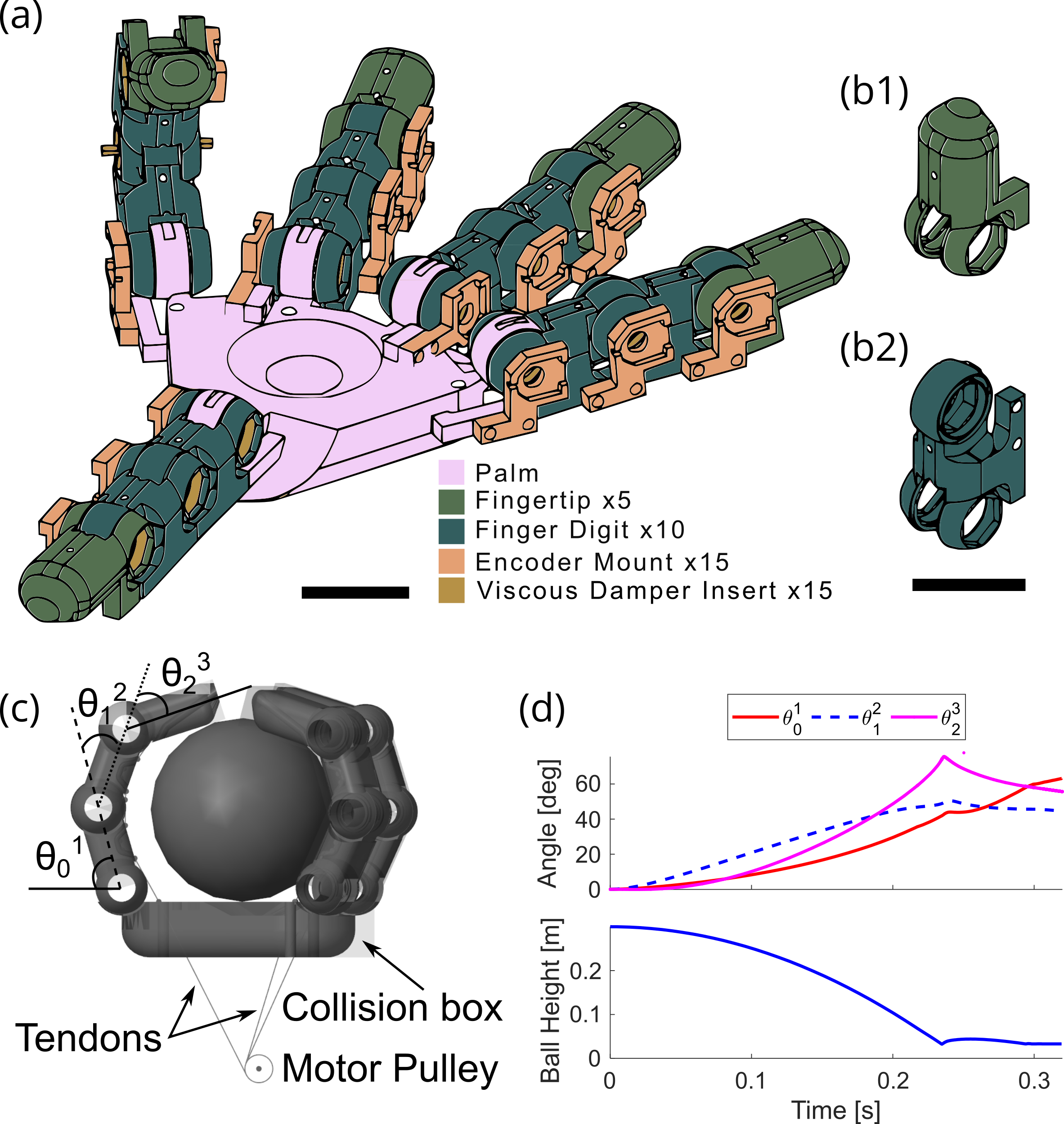}
    \caption{System Design Overview: (a) Isometric view of CAD assembly depicting 3D printed components. (b) CAD model of (b1) finger tip and (b2) finger digit. (c) Simulink Simscape model of hand and tendon system showing angular position \(\theta\) of articulated finger segments. (d) Mapping of articulated finger segment angle \(\theta\) with respect to ball height. All scale bars: 10mm.}
    \label{fig:system-design}
    \vspace{-25pt}
\end{figure}

We designed a bioinspired robotic hand capable of flexion and performing grasping tasks. The hand assembly (Fig. \ref{fig:system-design}(a)) is composed of five fingers attached around a central platform that acts as a palm. Each finger consists of two identical phalanges (Fig. \ref{fig:system-design}(b2)) and ends with a third phalange modeled as a fingertip (Fig. \ref{fig:system-design}(b1)). Each component in the assembly was designed to mimic the structure of the human finger, specifically concerning the length between pivot points and the relative size of each knuckle and phalange to the overall hand structure. Appropriate values were determined by measuring finger lengths from one of the authors. A concentric snap-fit feature joins each digit to form a hollow rotary knuckle into which the viscous damping joints can be inserted. The joint flexion is constrained with braces to prevent negative rotation past 180$^o$, and is driven by a Kevlar cable. The cable is routed through each phalange segment in series, akin to the flexor tendons of the human hand. The driving Kevlar cables of each finger pass through alignment holes in the palm before connecting to a series elastic element. This element consists of 24 parallel Nylon threads that achieve an equivalent stiffness of 9.52 N/mm to mimic the stiffness of human tendons\cite{Kamper2002}. The Nylon threads connect to a pulley on the actuating motor (Fig. \ref{fig:system-design}(c)). Activating the motor tensioned the cables, causing the fingers to contract. Another passive parallel elastic element made from a rectangular silicone strip is attached along the outside of each finger to replicate ligaments. Based on the geometry of the joint and the element and the material stiffness (shear modulus $\sim2.4$ kPa from tensile test, data not shown), the effective rotational stiffness of the element was estimated as $6.6\times10^{-2}$ Nm/rad. To enable measurement of individual joint rotations, each finger brace features a position encoder mount that is concentrically aligned with the axis of rotation. Clearances have been designed in each component fit to allow for low-friction rotation of the joints and avoid rubbing of the cable on the 3D-printed components. Elastic caps on the fingertips provide a non-slip contact region. 

\subsection{Modeling of Hand Dynamics}
A key aspect of the hand was the ability to respond quickly enough to move the fingers into position to capture the falling ball while also accounting for the inertial dynamics and viscoelastic forces at the joints. We created a Simulink Simscape (MATLAB, MathWorks) model of the hand and tendon system (Fig. \ref{fig:system-design}(c)) to calculate the net torque required to actuate the hand from the initial position (fingers in the palm plane) to the final position (fingers bent at 60$^o$ relative to each other) in 0.3 seconds given estimates of both passive and dynamic stiffness from previously recorded human data \cite{Kamper2002}. This actuation time was based on estimates of human reaction time in catching tasks \cite{Cesqui2016,owings_influence_2003}. Using the tool from Simscape's Multibody toolbox, we incorporated the dynamics of the fingers, the passive viscoelasticity of the joints, the routing and force transmission of the tendons, and contact interactions between the fingers and a grasped object. The geometry of the hand was imported from Solidworks, and the inertial properties and joint constraints were calculated. A series of small, negligible-weight pulleys simulated the routing of the tendon. Collision shapes were represented as rectangular bricks with the same inertial properties as the fingers.  

We utilized this model to determine the motor torque required to actuate the hand and catch a falling ball. With joint stiffness and damping equivalent to the levels observed in humans \cite{Kamper2002}, a torque of 0.4 Nm was sufficient to move the fingers in position to catch the ball in approximately 0.225 seconds (Fig. \ref{fig:system-design}(d)). Based on both the torque and speed requirements, we chose a Pololu 30:1 37D gear motor to drive the tendons of the robotic hand.



\subsection{Mechatronics System and Controller Design}
\begin{figure}[!b]
    \centering
    \vspace{-15pt}\includegraphics[width=\linewidth]{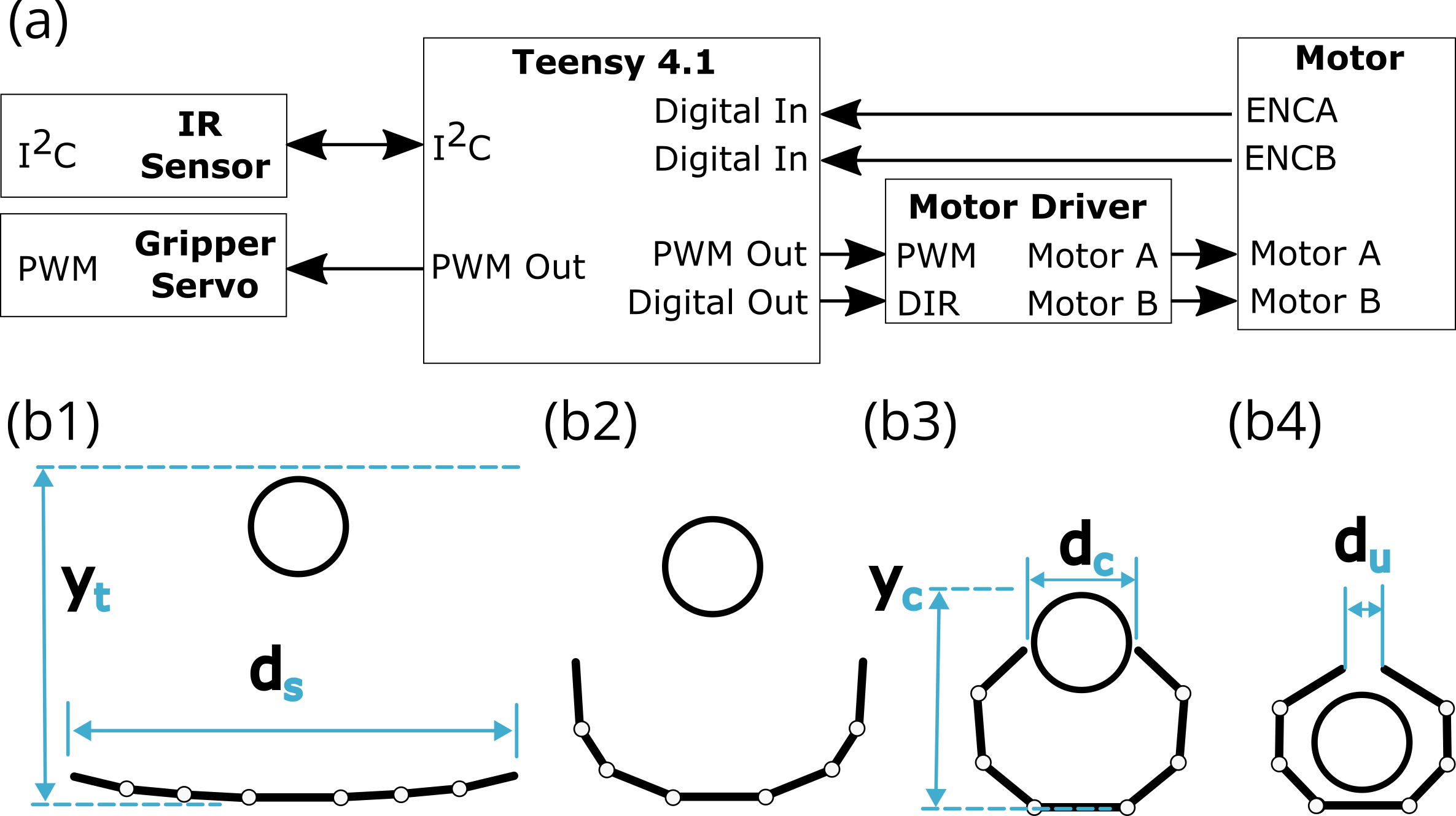}
    \caption{(a) Mechatronics system schematic. A Teensy 4.1 microcontroller was used to integrate signals from an IR distance sensor and motor encoders and to drive the tendon motor. (b) The Teensy implemented a closed-looped controller based on the height of the ball above the palm. }
    \label{fig:mech-sys}
\end{figure}

A brushed DC gear motor (Pololu 30:1 37D x 68L mm 12V Metal Gearmotor) with a 64 counts-per-revolution (CPR) encoder was used to drive the tendons via a 3D-printed 5 mm diameter pulley. The encoder was connected to a Teensy 4.1 to provide interrupt-driven updates of the angular position of the motor shaft. Motor speed control was executed via pulse width modulation (PWM) from a Teensy digital pin to a motor driver (Cytron MDD10A rev. 2). To track the position of the ball in real time, a time-of-flight laser distance sensor (Sparkfun VL53L1X module) was chosen for its accuracy ($\pm$ 5mm) over large ranges (4 cm to 4 m). To allow us to consistently and repeatably drop a ball into the hand to perform catch tests, the Teensy triggered a servomotor robotic claw (LewanSoul) to release the ball at a prescribed time. The laser distance sensor was mounted in the claw directly above the setup.

A closed-loop position controller was implemented to close the hand and capture the ball during drop tests. The controller used ball position data from the laser sensor and motor angle data from the encoder. For a drop test, the fingers start in an open position, with the distance between the fingertips equal to $d_s$ (Fig. \ref{fig:mech-sys}(b1)). While the ball's vertical position is still greater than a threshold ($y_t$), the hand is kept open. Below this threshold, the fingers' position is set as being linearly proportional to the distance of the ball to the hand (Fig. \ref{fig:mech-sys}(b2)). The proportionality constant is set such that when the ball reaches the capture height ($y_c$), the fingers are at a distance ($d_c$) (Fig. \ref{fig:mech-sys}(b3)). For ball heights less than $y_c$, the fingers are commanded to fully close around the ball at the final distance, $d_u$ (Fig. \ref{fig:mech-sys}(b4)). These parameters were experimentally tuned to improve catching performance. The angular positions of the fingertips and joints as a function of motor angle were empirically characterized  (Fig. \ref{fig:char-res}(b)) and used to determine the angular position-to-fingertip distance relationship. A low-level proportional controller with a gain of 20 was implemented to scale the PWM signal between 0 and 255 as a function of the error between the current and target position in degrees.

\subsection{Design and Modeling of Viscous Damping Components}

\begin{figure}[!t]
    \centering
    \includegraphics[width=\linewidth]{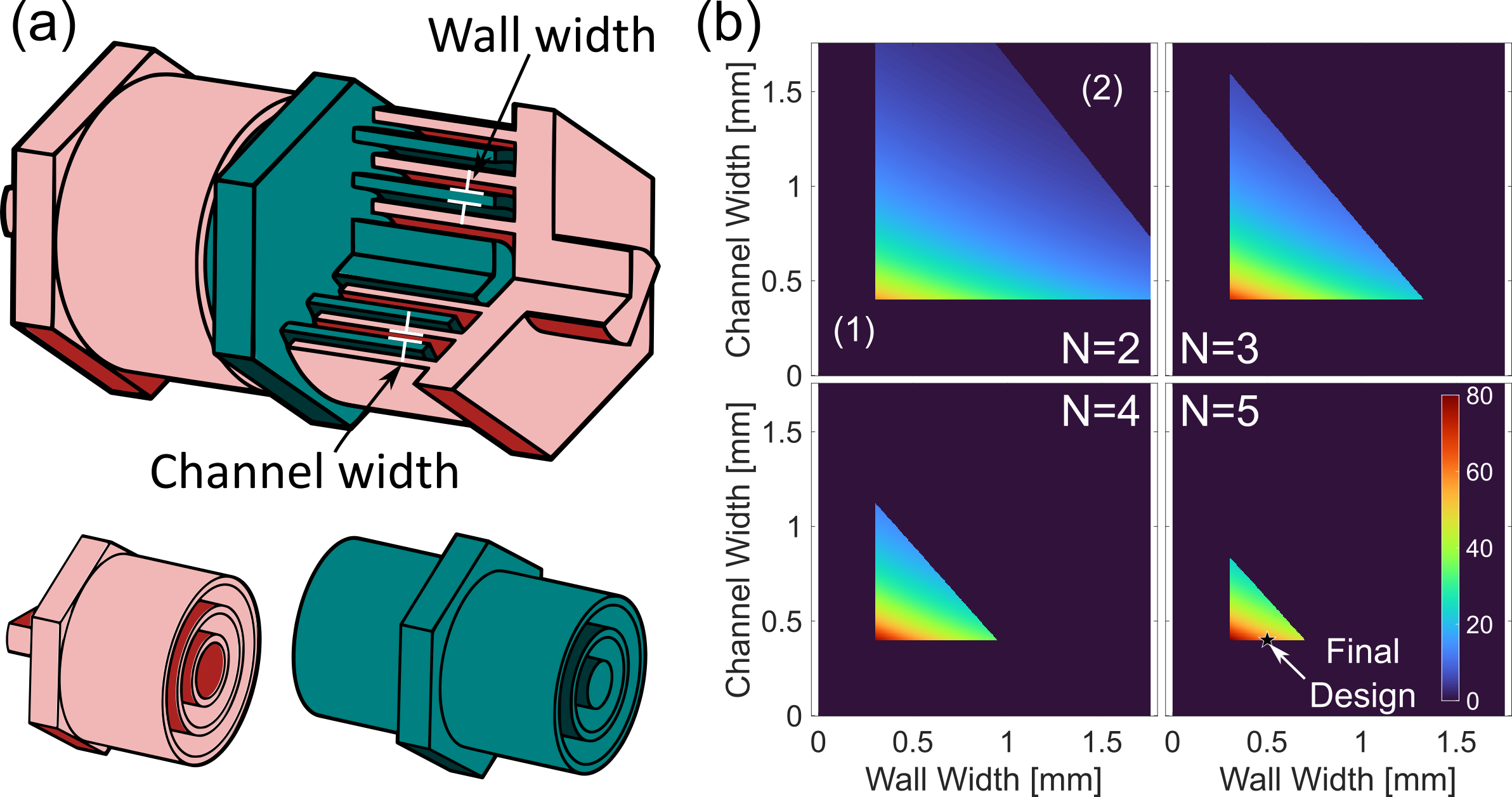}
    \caption{Viscous Damping Component Overview (a) CAD model and cross section of final concentric fin damper design. (b) Parametric sweep of normalized damping coefficient ($G$, reported in units $10^{-6}$ m$^3$). The $G$ factor is impacted by the channel width and wall width for a given number of internal fins ($N$). The region denoted (1) contains parameters below the printing tolerances of the printers used to fabricate the dampers, and the region denoted (2) indicates parameter pairs that would result in wall intersections. These regions exist for all panels. The same color map is used for all panels.}
    \label{fig:dyn_damp_model}
    \vspace{-20pt}
\end{figure}

To introduce viscous joint damping into the robotic finger joints, we designed a concentric ring damper utilizing principles of concentric cylinder rheometry (Fig. \ref{fig:dyn_damp_model}(a)). In this damper, torques are generated from fluid deformation between parallel surfaces and are directly proportional to the angular velocity of the joint. The dampers consist of inner and outer components. The inner and outer components interface with the proximal and distal phalanges, respectively, via a hexagon key. These components contain a series of concentric cylindrical fins that interdigitate with each other when the damper is assembled. The cylinders are coaxial with the axis of joint rotation. The void between the fins is filled with a working fluid, and as the joint rotates, the components move through the fluid, generating shear stress in the fluid, which reacts on the solid components and generates a resisting torque.

We developed a model based on Couette flow \cite{Illingworth1950} to determine the relationship between the joint rotation, the working fluid's viscosity, the damper geometry, and the torques generated by the damper. This model assumes steady-state circumferential flow of a Newtonian fluid. The velocity gradient between interdigitated elements is found from the geometry and angular velocity and is multiplied by viscosity to calculate the shear stress. Finally, the shear stress is multiplied by the moment arm and integrated across the parallel surfaces to calculate the resulting torques. Because of the parameter linearity in the Navier-Stokes equations, this torque equation can be written as $T = -\mu G \dot\theta$, where $T$ is the torque, $\mu$ is the working fluid viscosity, $\dot\theta$ is the joint angular velocity, and $G$ is a factor accounting for the geometry of the damper (see Appendix \ref{section:ViscDamp}). $G$ can be considered a viscosity-normalized damping coefficient and is parametrically defined by the free geometric parameters of the damper.  


Parametric sweeps of the free geometric parameters (channel width and cylinder wall width) were performed to determine the level of damping that could be achieved by the damper (Fig. \ref{fig:dyn_damp_model}(b)). The number of fins, the width of the fin, and the width of the channel are restricted by the joint-bound geometry. These parametric sweeps allowed us to determine the viscosity of the working fluid required to achieve near-human levels of joint damping ($[8.1,14.2]\times10^{-3}$ N m s/rad \cite{Kamper2002}) based on various fin geometries. The optimal design that was feasible to fabricate consisted of five fins with a 0.5 mm wall width, 0.4 mm channel width, and a required working fluid viscosity of [135,000, 236,000] cP. In an effort to approach this high viscosity range and comply with our motivation of using accessible and organic materials, we selected peanut butter (viscosity commonly reported by industry resources as 150,000-250,000 cP \cite{pb-viscosity, viscosityScaleGuide}) as the working fluid to be used for all fabricated dampers. 
\subsection{Design and Model Availability}
The complete design of the hand and its subsystems, as well as all model code, is available on GitHub (\href{https://github.com/CMU-BORG/PeanutButter-and-Joints}{GitHub/PeanutButter-and-Joints}). An archived version can be found on Zenodo (doi: 10.5281/zenodo.15257954).

\section{Fabrication and Characterization Methods}

\subsection{3D Printing of Hand and Damper Components}
All digital models in this study were created in Solidworks 2022 (Dassault Systèmes). Geometric codes (g-code) of the fingers and palm were generated using PrusaSlicer v2.5.1. These components were printed with PLA using a 0.2 mm resolution on a Prusa Original MK3 printer. Components of the viscous dampers were sliced using Lychee 5.0 (Mango 3D) resin slicer and printed in Aqua-Gray 8K resin on a Sonic Mini 8K printer (Phrozen Technology). Printed parts were washed with isopropanol in an ultrasonic cleaner (Quantrex 90H, L\&R Ultrasonics), and post-cured in the Phrozen Cure station for 10 minutes. 
\vspace{-15pt}
\subsection{Fabrication of Elastic Components}
All elastic components were fabricated using DragonSkin 30 (Smooth-On Inc). Per manufacturer's instructions, equal parts by weight of precursors A and B were mixed thoroughly and degassed in a vacuum chamber for approximately 5 minutes. Ligament elements were cast in a laser-cut acrylic mold (internal dimensions 100mm x 5.2 mm x 2.6mm). Finger caps were cast in a 3D-printed (Prusa Original MK3), multi-component mold sprayed with mold release (Ease Release 200, Mann Release Technologies). All components were cured at room temperature overnight. The ligament analogs were fixed to the back of each finger using a Clear Adhesive Sealant Silicone RTV 80050 (Permatex). A small amount of product was applied between the back of each digit and the Dragonskin and manually clamped for 20 minutes. The entire hand cured for 24 before use. Finger caps were placed over the fingers and secured with mechanical retention.
\vspace{-15pt}

\subsection{Concurrent Flexion Characterization}
The mapping of fingertip position and motor angles were empirically obtained for fingers with and without the elastic element. High-contrast markers were placed on each of the three joints and the fingertip. The motor was swept from 0$^\circ$ to 270$^\circ$ in 10$^\circ$ increments, with 2 seconds between each move to allow the position to stabilize before moving again. Finger motion was recorded with a camera (60 FPS) and markers were tracked using Tracker 5.1.5 (physlets.org/tracker). To validate concurrent flexion, measured joint angles and motor position were normalized between $[0,1]$. Within each test condition (with and without elastic element), we calculate all pairwise correlation coefficients between joint kinematic time series. High correlation coefficients (near 1) indicate concurrence in flexion.

\vspace{-15pt}

\subsection{Pendulum Test Damper Characterization}
To determine the frictional properties of the joint running surface and the damping coefficient of the fabricated viscous dampers, pendulum tests were conducted. The pendulum joint was a modified version of the finger joints in which each distal edge of the component was modified to hold a rectangular acrylic bar coplanar with the mid-joint plane (Fig. \ref{fig:char-res}(c)). One bar was vertically fixed, and the free end was weighted with a nut, bolt, and washers (11.2 g). A viscous damper was inserted into the joint for damping parameter tests, but the joint was left empty for friction property tests. For each test condition, 15 drop tests were conducted in which the free end of the pendulum was manually raised, dropped, and allowed to oscillate until coming to rest. The pendulum motion was recorded using an iPhone 6s camera in SLO-MO mode (Apple Inc.; 240 FPS, 720 p). The position of the pendulum weight was tracked using Tracker 5.1.5 (physlets.org/tracker), and the angular position was calculated. 

An analytical model of the pendulum dynamics (Fig. \ref{fig:char-res}(c)) was constructed accounting for Coulomb and linear viscous friction in the joint running surface, and for viscous torques generated by the damper (Appendix \ref{section:Pend}). Dynamic frictional coefficients were fit to the damper-free results, and the damping coefficient was fit to the damped results. Parameter distributions were determined using a bootstrapped parameter estimation routine. The details of this routine can be found with the code in our repository. 




\section{Results and Discussion}

\subsection{Effects of Passive Elastic Elements on Hand Flexion}
\begin{figure}[!tbp]
    \centering
    \includegraphics[width=0.9\linewidth]{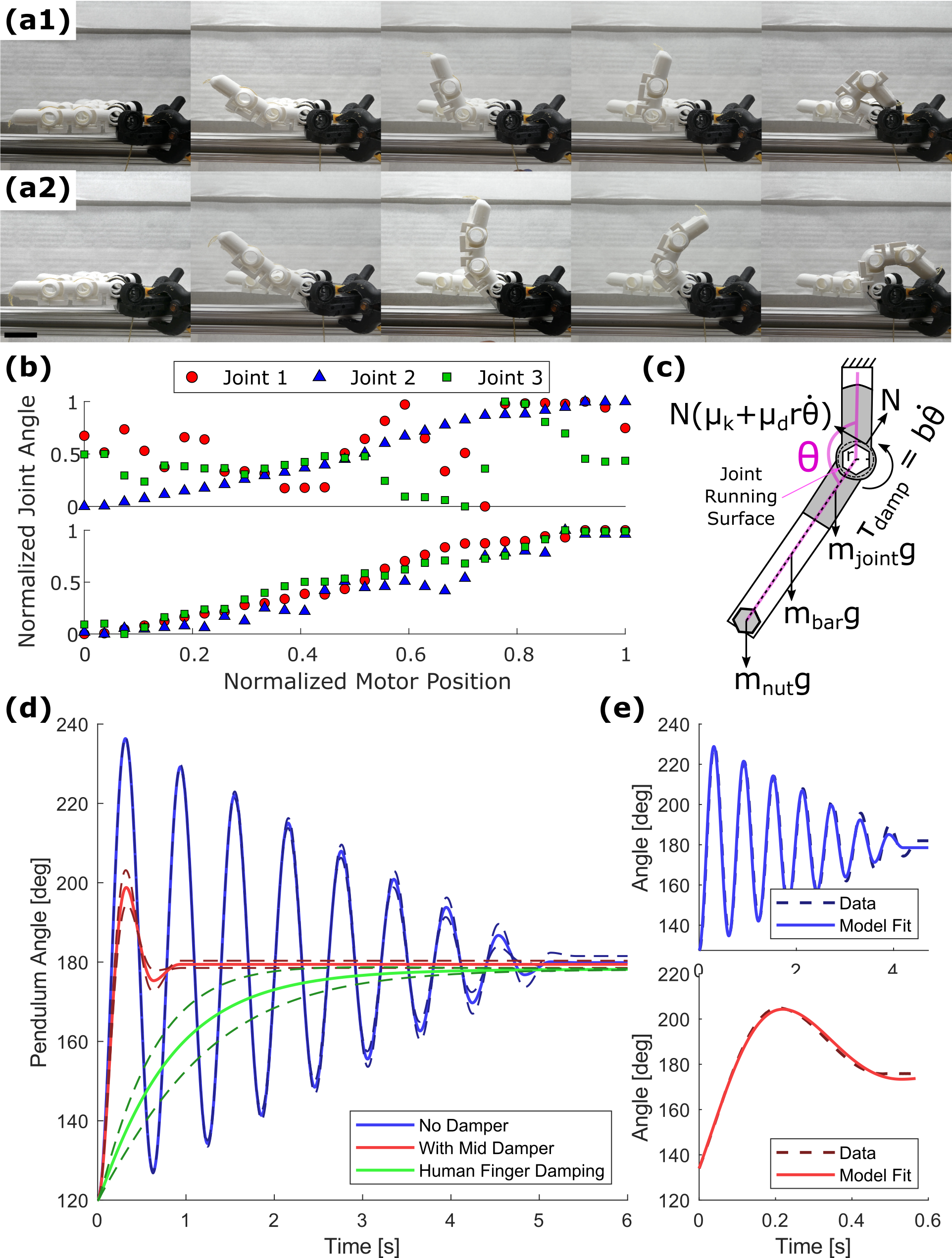}
    \vspace{-10pt}
    \caption{Experimental characterization of robotic subsystems. (a) Robot finger behavior during contraction without (a1) and with (a2) an elastic element. Concurrent flexion is only observed with a parallel elastic element (Scale bar: 30mm). (b) Range-of-motion normalized time series of the finger joint angles throughout the motor range of motion. Top: no elastic element. Bottom: elastic element present. Concurrent flexion is shown in the high correlation between the time series of the three different joints in the elastic case. (c) Force diagram for pendulum drop experiments. (d) Monte Carlo mean (solid) and 95\% confidence interval (dashed) dynamics of a finger digit pendulum for undamped, PB damped, human damped cases. Parameters for the No Damper and Damper cases were drawn from the joint parameter distribution obtained from our bootstrapping estimation routine (N=20). Human damping values were drawn from a uniform distribution ($b_{human}\sim U([8.1,14.2]\times10^{-3})$). (e) Example fits of the dynamical model fit vs pendulum data for the undamped (top) and PB damped (bottom) case.}
    \label{fig:char-res}
\end{figure}

The addition of a passive elastic element to the fingers was vital to achieving concurrent flexion and, therefore, critical to the successful completion of grasping tasks. Initially, we observed the behavior of the tendon-driven fingers during contraction without the presence of an elastic element in parallel (Fig. \ref{fig:char-res}(a1)). When actuated, the finger articulates starting from the distal joints first and only demonstrates articulation in the more proximal joints once the former has maximized its rotation. The relative angle of rotation in each joint for this case can be seen in Fig. \ref{fig:char-res}(b). However, when the elastic ligament analog element was introduced to the back of the fingers, all three joints rotated simultaneously (Fig. \ref{fig:char-res}(a2,b)). This translates to concurrent flexion of the entire finger and results in more organic movement that enables the fingers to grasp around an object. This is quantitatively supported by the pair-wise correlation coefficients obtained for each group. With no elastic element, the average pair-wise correlation was 0.350 (range: [0.220,0.443]), whereas with an elastic element, the average was 0.959 (range: [0.945,0.972]). The much higher correlation in the finger with elastic elements supports the idea that passive elastic elements assist in synchronizing joint flexion. Additionally, the correlations in the elastic group being near 1 suggest that the finger is successfully concurrently flexing.



\vspace{-15pt}

\subsection{Effects of Viscous Dampers on Joint Dynamics}
The effect of our viscous dampers on the passive dynamics of the finger joints was determined using pendulum drop tests. In control tests, a pendulum (as described above and shown in Fig. \ref{fig:char-res}(c)) with no damper at the joint was allowed to oscillate, coming to rest only under the influence of friction. In damped tests, a damper was inserted into the same pendulum, and the drop tests were repeated. The pendulum showed a very underdamped response in the control cases, oscillating $8.0\pm1.0$ times, coming to rest after $4.9\pm0.5$ s. Conversely, in damped cases, the pendulum always came to rest after just 1 oscillation, in $0.48\pm0.04$ s. 

The angle measured during these experiments fit well with the presented model (Fig. \ref{fig:char-res}(e)) in both the undamped and damped cases. Using a bootstrapped parameter estimation method, we obtained estimates for the coefficients of kinetic friction (mean [95\% Credible Interval]: $2.88\text{ } [2.66,3.10]\times10^{-3}$), as well as the damping coefficient for the torsional damper ($0.759\text{ } [0.671,0.920] \times10^{-3}$ N m s/rad (mean [95\% Credible Interval)). The coefficient of viscous friction was much smaller than kinetic friction ($0.0013\text{ } [0.0000,0.0118] \times10^{-3}$) and heavily right-skewed (median: $2.8\times10^{-9}$). This suggests that this parameter is non-zero predominantly because of the penalty terms in the optimization algorithm and does not play a significant role in the dynamics. 

Finally, the mean and 95\% confidence intervals on the undamped and damped pendulum dynamics were calculated using the bootstrapped parameter distributions. The damped case was calculated for both the level of damping we achieved as well as the human level of damping. In contrast to our parameter sweep predictions, we were not able to achieve human-level damping, which would cause an overdamped response in the pendulum. However, we were able to reduce the settling time of the joints dramatically compared to the undamped case, and with further refinement, we may better match the level of damping seen in human joints. 

The level of damping observed in the fabricated dampers was much lower than predicted by the analytical model (approximately $9 \text{ to }15\times10^{-3}$ N m s/rad from the model compared to $0.758\text{ } [0.670,0.939] \times10^{-3}$ N m s/rad that was achieved). Many reasons in both the modeling and experimentation could cause this discrepancy. Within the model, the fluid was assumed to be at steady state, while in reality, much of the fluid motion will be non-steady state. However, this would likely cause the model to underpredict the damping. Additionally, the working fluid was assumed to be Newtonian, while peanut butter is better described as a thixotropic Bingham plastic. Thus, while the apparent viscosity is very high at rest, the speed of rotation in the pendulum tests may apply sufficient shear rates to decrease the apparent viscosity below the 150,000-250,000 cP range \cite{Citerne2001}. Likely, the main reason for the underperformance comes instead from the fabrication. The small geometry made it difficult to consistently create an even thin film of peanut butter on all fin surfaces. When placing the dampers into the joints, there was some squeeze out that could not back-flow into the damper, causing voids in the peanut butter. Additionally, the very thin layer of peanut butter ($\sim$400 $\mu$m) was so thin and viscous that it was likely not acting as a continuous film but as a series of discontinuous regions of lubrication. 

\subsection{Demonstration of Successful Ball Catching}
\begin{figure}[!t]
    \centering
    \includegraphics[width=\linewidth]{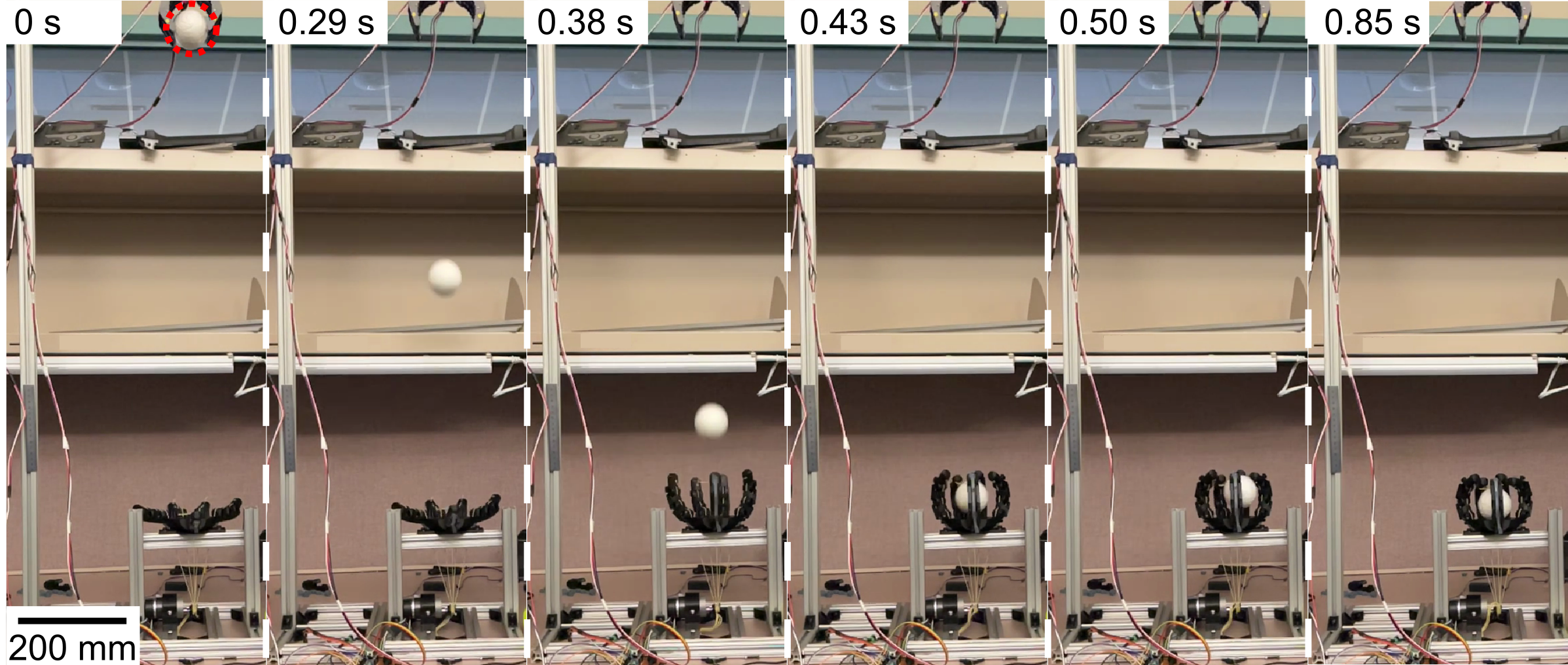}
    \caption{Sequential frames demonstrating a successful ball catch by the robot hand. The ball is highlighted by a red dashed circle in the first frame.}
    \label{fig:catch}
\end{figure}
When experimentally tested, the robotic hand was capable of repeatedly catching a falling ball (Fig. \ref{fig:catch}). The implementation of the closed-loop controller accurately mapped the position of the falling ball and triggered the actuation of the fingers in time to intercept. Concurrent flexion of the joints facilitated the wrapping of the fingers around the ball once to arrest its motion and secure it in the palm.  Out of $N=22$ ball drop trials, the hand had a successful catch rate of 59\%. 

\section{Conclusions and Future Work}
In this work, we presented the design, modeling, fabrication, and testing of an open-source bioinspired grasper that incorporates morphological and mechanical elements of the human hand. We demonstrated that for an underactuated hand with articulated segments, the passive mechanical properties of elastic elements mimicking tendons and ligaments are of critical importance to induce concurrent flexion. Furthermore, we prototyped modular viscous damper inserts that successfully damped the response of the articulated finger segments using accessible materials. The robotic system was able to successfully perform catching tasks, even when using a simple closed-loop controller to map the position of a falling ball. In the design and fabrication of our biomimetic hand robot, we prioritized the use of low-cost, accessible materials, including the choice of an organic and biocompatible working fluid (peanut butter).  

Future work will iterate upon the proposed viscous damper design to achieve human levels of damping. Additional modifications to the hand could be made to improve its robustness, such as increasing the strength of the tendon and ligament analogs to avoid breaking from violent impacts, and more efficient routing of the tendons for torque optimization. The presented platform will enable us to experimentally explore the efficacy of joint damping in absorbing impulsive impacts. We hope to pursue these improvements to allow us to test questions \textit{in roboto} that would be difficult to test \textit{in vivo} or \textit{in silico}, such as how these passive viscoelastic elements help to compensate for bandwidth limitations in the nervous system to perform dynamic and rapid manipulation tasks.

\section*{Acknowledgment}
The authors would like to thank Nolen Keeys, the teaching team, and CMU MechE department funding in support of 24-775: Bioinspired Design and Experimentation. We also thank Professor Nancy Pollard for her helpful guidance on simulating the role of damping in the dynamics of grasping. This work was supported in part by the National Science Foundation (NSF) through grant no. FRR-2138923. Any opinions, findings, and conclusions expressed in this material are those of the authors and do not necessarily reflect the views of the NSF.

\section{Appendices}
\appendix

\section{Modeling of Concentric Ring Dampers}\label{section:ViscDamp}

The G factor in the damper torque equation can be estimated from the geometry of the damper. Here, we model the dampers using Couette flow (steady-state, circumferential flow only, and rotational symmetry). For the concentric fin damper, each fin in the concentric cylinder will contribute to the total torque applied to the joint in three ways: 1) shear stresses from the fluid between the fin and its immediately medial neighbor, 2) shear stresses from the fluid between the fin and its immediately lateral neighbor, and 3) shear stress from the fluid between the end of fin and the base of the opposite portion of the damper. For a fin whose midline is a distance $\rho$ from the center of the damper with a wall thickness w and length L, separated from the next fin by a channel of width $\delta$, the shear stress from the fluid on the medial and lateral surfaces is:
\begin{equation}
    \tau = -\mu\omega\frac{(\rho \pm w/2 \pm\delta)^2}{\delta(2(\rho\pm w/2)\pm\delta)}
\end{equation}
($+$: lateral, $-$: medial). The torque from this stress is calculated by multiplying by the moment arm and integrating over the surface area, yielding:
\begin{equation}
    T = -4\pi\mu\omega \frac{(\rho \pm w/2 \pm \delta)^2(\rho\pm w/2)^2L}{\delta(2(\rho\pm w/2)\pm\delta)}
\end{equation}
The torque on the fin end is found similarly by multiplying the shear stress by the moment arm and integrating over the surface. This yields:
\begin{equation}
    T_{end} = -\frac{\pi\mu\omega}{2\delta}\left((\rho + w/2)^4 - (\rho - w/2)^4 \right)
\end{equation}
Factoring out the term of $\mu\omega$ gives the G factor for a full internal fin as:
\begin{equation}
    \begin{split}
        G_i(\rho_i,w,\delta) = 4\pi\biggl[& \frac{(\rho_i - w/2 - \delta)^2(\rho-w/2)^2L}{\delta(2(\rho_i-w/2)-\delta)}  + \frac{(\rho_i + w/2 + \delta)^2(\rho_i+w/2)^2L}{\delta(2(\rho_i+w/2)+\delta)} \\ +  &\frac{\left((\rho_i + w/2)^4 - (\rho_i - w/2)^4 \right)}{8\delta} \biggr]
    \end{split} \end{equation}
For the inner pin, the first term drops out, and for the outermost fin, the middle term drops out. Then the total G factor is found by adding together all of the G factors for the fins on one half of the damper (only half to avoid double-counting).

\section{Modeling of Pendulum Dynamics}\label{section:Pend}
The dynamics of the pendulum were derived following the free-body diagram found in Fig. \ref{fig:char-res}(c). Three coupled bodies were considered: the 3D-printed lower joint, the lower acrylic bar, and the bolt-nut-washer weight. The system has one degree of freedom -- the angle from the positive vertical axis to the midline of the lower pendulum arm. The weight was modeled as a point mass, while the other components were modeled as rigid bodies. The lengths used in the model and the moment of inertia of the joint ($I_{joint}$) were measured from the CAD assembly in Solidworks. The moment of inertia of the bar was calculated assuming the bar was a perfect rectangular prism rotated about a parallel axis 
\begin{equation}
    I_{bar} = (1/12)m_{bar}(L_{bar}^2 + w_{bar}^2) + m_{bar}r_{bar}^2
\end{equation}
where $m_{bar}$ is the mass of the bar (measured), $L_{bar}$ is the length of the bar, $w_{bar}$ is the width of the bar in the plane, and $r_{bar}$ is the distance from the axis of rotation to the center of mass of the bar). The angular inertia of the point mass is $I_{weight}=m_{weight}r_{weight}^2$, where $m_{weight}$ and $r_{weight}$ are the mass and distance from the rotation axis to the weight. The total moment of inertia of the system is the sum of the component moments of inertia.

The torques in the system arise from the vertical components of the gravitational forces acting on each body at that body's centers of mass, frictional forces acting at a point on the running surface in line with the central axis, and the torque generated by the viscous damper ($\tau_{damp} = b\dot\theta$). The frictional forces were assumed to be a combination of Coulomb kinetic friction and viscous friction: 
\begin{equation}
    F_{fric} = -\text{sign}(v) N(\mu_k + \mu_d v)
\end{equation}
where $N$ is the normal force between the contacting edge of the running surface, $\mu_k$ and $\mu_d$ are the coefficients of kinetic and viscous friction, respectively, and $v$ is the tangential velocity of the running surface at the contact point. Here, $v=r_{joint}\dot\theta$, where $r_{joint}$ is the radius of the joint running surface. The normal force $N$ comes from a combination of the centripetal force and the projection of gravity along the length of the pendulum:
$$
N = \left(\sum{m_i r_i}\right)\dot\theta^2  + \left(\sum{m_i r_i}\right)g\cos(\theta-\pi)
$$
This yielded the equation of motion:
\begin{equation}
    \ddot\theta = \frac{1}{I}\left[\left(\sum{m_i r_i}\right)g\sin(\theta) - b\dot\theta \right.\left.- Nr_{joint}\text{sign}(\dot\theta)\left(\mu_k + \mu_dr_{joint}\dot\theta\right)  \right]
\end{equation}
This equation was integrated using a forward explicit Euler method with a timestep set at 1\% of the experimental timestep (experimental $\Delta t$ = 1/240 s).

\def\url#1{}
\bibliographystyle{splncs04}
\bibliography{refs}

\begin{thebibliography}{10}
\providecommand{\url}[1]{\texttt{#1}}
\providecommand{\urlprefix}{URL }
\providecommand{\doi}[1]{https://doi.org/#1}

\bibitem{Guizzo}
By leaps and bounds an exclusive look at how boston dynamics is redefining robot agility by erico guizzo • photography by bob o'connor, \url{https://spectrum.ieee.org/bostondynamics1219}

\bibitem{Fundamentals}
Fundamentals of Sensory Physiology. Springer Berlin Heidelberg (1981). \doi{10.1007/978-3-662-01128-7}

\bibitem{Burridge}
Burridge, R.R., Rizzi, A.A., Koditschek, D.E.: Sequential composition of dynamically dexterous robot behaviors

\bibitem{Cesqui2016}
Cesqui, B., Russo, M., Lacquaniti, F., D'Avella, A.: Grasping in one-handed catching in relation to performance. PLoS ONE  \textbf{11} (7 2016). \doi{10.1371/journal.pone.0158606}

\bibitem{Citerne2001}
Citerne, G.P., Carreau, P.J., Moan, M.: {Rheological properties of peanut butter}. Rheologica Acta  \textbf{40}(1),  86--96 (2001). \doi{10.1007/s003970000120}

\bibitem{Dong2021}
Dong, R.G., Wu, J.Z., Xu, X.S., Welcome, D.E., Krajnak, K.: A review of hand–arm vibration studies conducted by us niosh since 2000. Vibration  \textbf{4},  482--528 (6 2021). \doi{10.3390/vibration4020030}

\bibitem{guizzo_darpa_nodate}
Guizzo, E., Ackerman, E.: {DARPA} {Robotics} {Challenge}: {A} {Compilation} of {Robots} {Falling} {Down} - {IEEE} {Spectrum}, \url{https://spectrum.ieee.org/darpa-robotics-challenge-robots-falling}

\bibitem{holme_nielsen_rapid_2018}
Holme~Nielsen, C., Bladt~Brandt, A., Thymann, T., Obelitz-Ryom, K., Jiang, P., Vanden~Hole, C., van Ginneken, C., Pankratova, S., Sangild, P.T.: Rapid {Postnatal} {Adaptation} of {Neurodevelopment} in {Pigs} {Born} {Late} {Preterm}. Developmental Neuroscience  \textbf{40}(5-6),  586--600 (2018). \doi{10.1159/000499127}

\bibitem{Illingworth1950}
Illingworth, C.R.: Some solutions of the equations of flow of a viscous compressible fluid. Mathematical Proceedings of the Cambridge Philosophical Society  \textbf{46},  469--478 (1950). \doi{10.1017/S0305004100025986}

\bibitem{Ingram2008}
Ingram, J.N., Körding, K.P., Howard, I.S., Wolpert, D.M.: The statistics of natural hand movements. Experimental Brain Research  \textbf{188},  223--236 (6 2008). \doi{10.1007/s00221-008-1355-3}

\bibitem{Kamper2002}
Kamper, D.G., Hornby, T.G., Rymer, W.Z.: Extrinsic flexor muscles generate concurrent flexion of all three finger joints (2002)

\bibitem{Lakie2012}
Lakie, M., Vernooij, C.A., Osborne, T.M., Reynolds, R.F.: The resonant component of human physiological hand tremor is altered by slow voluntary movements. Journal of Physiology  \textbf{590},  2471--2483 (5 2012). \doi{10.1113/jphysiol.2011.226449}

\bibitem{Liu2007}
Liu, D., Todorov, E.: Evidence for the flexible sensorimotor strategies predicted by optimal feedback control. Journal of Neuroscience  \textbf{27},  9354--9368 (8 2007). \doi{10.1523/JNEUROSCI.1110-06.2007}

\bibitem{Murray1994}
Murray, R.M., Li, Z., Sastry, S.: A mathematical introduction to robotic manipulation. CRC Press (1994)

\bibitem{Namiki2003}
Namiki, A., Imai, Y., Ishikawa, M., Kaneko, M.: Development of a high-speed multifingered hand system and its application to catching. vol.~3, pp. 2666--2671 (2003). \doi{10.1109/iros.2003.1249273}

\bibitem{Tuong}
Nguyen-Tuong, D., Peters, J.: Model learning for robot control: A survey (11 2011). \doi{10.1007/s10339-011-0404-1}

\bibitem{nishikawa_neuromechanics_2007}
Nishikawa, K., Biewener, A.A., Aerts, P., Ahn, A.N., Chiel, H.J., Daley, M.A., Daniel, T.L., Full, R.J., Hale, M.E., Hedrick, T.L., Lappin, A.K., Nichols, T.R., Quinn, R.D., Satterlie, R.A., Szymik, B.: Neuromechanics: an integrative approach for understanding motor control. Integrative and Comparative Biology  \textbf{47}(1),  16--54 (Jul 2007). \doi{10.1093/icb/icm024}, \url{https://doi.org/10.1093/icb/icm024}

\bibitem{owings_influence_2003}
Owings, T.M., Lancianese, S.L., Lampe, E.M., Grabiner, M.D.: Influence of {Ball} {Velocity}, {Attention}, and {Age} on {Response} {Time} for a {Simulated} {Catch}. Medicine \& Science in Sports \& Exercise  \textbf{35}(8), ~1397 (Aug 2003). \doi{10.1249/01.MSS.0000078926.53402.9C}

\bibitem{pb-viscosity}
Scientific, P.: Viscosity of peanut butter cream (2018), \url{https://prime.erpnext.com/blog/application-knowledge/viscosity-of-peanut-butter-cream}, accessed: 2023-04-17

\bibitem{viscosityScaleGuide}
Smooth-On, I.: Viscosity scale reference guide, \url{https://www.smooth-on.com/assets/pdf/Viscosity_Scale_Reference_Guide.pdf}, accessed: 2023-04-17

\bibitem{Sutton}
Sutton, G., Szczecinski, N., Quinn, R., Chiel, H.: Neural control of rhythmic limb motion is shaped by size and speed

\bibitem{Vernooij2015}
Vernooij, C.A., Lakie, M., Reynolds, R.F.: The complete frequency spectrum of physiological tremor can be recreated by broadband mechanical or electrical drive. Journal of Neurophysiology  \textbf{113},  647--656 (1 2015). \doi{10.1152/jn.00519.2014}

\end{thebibliography}

\end{document}